%% file: main.tex
\ifdefined\pdfminorversion
\fi

\documentclass{article}
\usepackage{oc_opd_arxiv,times}
\iclrfinalcopy

\usepackage[utf8]{inputenc}
\usepackage[T1]{fontenc}
\usepackage{microtype}
\usepackage{graphicx}
\usepackage{booktabs}
\usepackage{array}
\usepackage{dcolumn}
\usepackage{amsmath,amssymb,amsfonts}
\usepackage{mathtools}
\usepackage{enumitem}
\usepackage{xcolor}
\usepackage{placeins}
\usepackage{hyperref}
\hypersetup{hidelinks}
\usepackage{cleveref}

\graphicspath{{figures/}}
\newcolumntype{d}[1]{D{.}{.}{#1}}

\newcommand{\figuremock}[3]{%
  \begingroup
  \setlength{\fboxsep}{0pt}%
  \fcolorbox{gray!45}{gray!4}{%
    \parbox[c][#1][c]{0.96\textwidth}{%
      \centering
      \color{gray!70}
      \textbf{#2 placeholder}\\[4pt]
      \small Final vector asset pending\\
      \texttt{#3}%
    }%
  }%
  \endgroup
}

\title{Outcome-Confounded Local Supervision in On-Policy Distillation}
\author{\input{author_arxiv}}

\begin{document}
\maketitle
\lhead{Preprint}

\begin{abstract}
\input{sections/00_abstract}
\end{abstract}

\input{sections/01_introduction}
\input{sections/02_background}
\input{sections/03_diagnostic}
\input{sections/04_discovery}
\input{sections/05_robustness}
\input{sections/06_training_dynamics}
\input{sections/07_exploratory}
\input{sections/08_related_work}
\input{sections/09_limitations_conclusion}

\bibliography{refs}
\bibliographystyle{oc_opd_arxiv}

\appendix
\input{sections/a_appendix}

\end{document}

%% file: author_arxiv.tex
Guoqing Ma\\
The Chinese University of Hong Kong, Shenzhen\\
\texttt{magq9807@gmail.com}

%% file: sections/00_abstract.tex
On-policy distillation (OPD) trains a student on its own trajectories while a
teacher supplies dense token-level likelihoods at student-visited prefixes.
These likelihoods are often read locally: agreement appears safe to imitate,
whereas disagreement appears to identify an error.  We show that both readings
are confounded by the outcome of the completed trajectory.  We introduce an
outcome-resolved diagnostic that crosses pointwise teacher--student divergence
with final-answer correctness, separating safe imitation, productive
divergence, harmful divergence, and agreement-on-failure.  In an eight-seed
mathematical-reasoning study with a Qwen3-8B student and Qwen3-32B teacher,
agreement-on-failure constitutes \(67.84\%\) of pooled response-token mass;
with a Qwen2.5-7B/32B pair it remains \(67.68\%\).  The result persists across
threshold, sequence-level, format, and truncation audits.  Even on prompts that
the Qwen3 teacher solves in all four independent attempts, student accuracy
rises to \(86.91\%\) but agreement-on-failure remains \(14.76\%\).  We then run
three matched training probes that use the available signals to imitate, mask,
or contrast whole trajectories; none consistently reduces
agreement-on-failure.  The result points to a localization limitation: local
divergence paired with a trajectory-level outcome does not identify where a
failed trajectory became unrecoverable.  Addressing this limitation requires
additional positional information, such as process labels, teacher
continuations from student prefixes, or token-level alignment across rollouts.
Our contribution is therefore diagnostic rather than a new training method.

%% file: sections/01_introduction.tex
\section{Introduction}

On-policy distillation (OPD) combines two appealing properties for reasoning
post-training.  The student generates the trajectories on which it is trained,
reducing the prefix mismatch induced by distillation on fixed teacher traces,
while a larger teacher provides dense token-level supervision at the states
the student actually visits \citep{agarwal2024opd}.  Relative to a sparse
final-answer reward, this appears to offer substantially richer feedback:
teacher and student probabilities can be compared at every response token.

The meaning of that comparison is often inferred locally.  A sampled token
that receives high teacher likelihood appears safe to imitate; a token on
which teacher and student differ appears to identify a mistake or correction
opportunity.  Reasoning success, however, is defined over the completed
trajectory.  After an early mistake, many continuations can remain locally
plausible while the response still ends incorrectly.  Conversely, a correct
response can depart from the teacher through an alternative derivation,
notation, or intermediate choice.  Local agreement is therefore not process
correctness, and local disagreement is not automatically error.

We call this mismatch \emph{outcome-confounded local supervision}.  As
illustrated in \cref{fig:diagnostic_lens}, our diagnostic crosses the local
teacher--student relation with the final outcome of the same student
trajectory.  Low divergence on a correct response is \emph{safe imitation};
high divergence on a correct response is \emph{productive divergence}; high
divergence on a wrong response is \emph{harmful divergence}; and low
divergence on a wrong response is \emph{agreement-on-failure}.  The four cells
separate two questions that a single local divergence score conflates:
whether the student locally follows the teacher and whether the complete
trajectory succeeds.

In controlled mathematical-reasoning studies, this decomposition exposes a
large residual failure regime.  Across eight discovery seeds with a Qwen3-8B
student and Qwen3-32B teacher, agreement-on-failure accounts for \(67.84\%\) of
pooled valid response-token mass at the default threshold.  Repeating the same
diagnostic with a Qwen2.5-7B-Instruct student and Qwen2.5-32B-Instruct teacher
gives the same qualitative pattern, with agreement-on-failure at \(67.68\%\) of
token mass.  The dominant region is therefore not conspicuous disagreement, but
low pointwise divergence on responses that ultimately fail.  Productive and
harmful divergence are both present, so uniformly suppressing or encouraging
divergence would also merge training cases with opposite outcomes.

We test several alternative explanations.  Agreement-on-failure remains large
over a five-point threshold sweep.  It is not produced solely by a few long
wrong responses: in three-seed sequence-level reruns,
\(89.15\%\)--\(100\%\) of wrong trajectories contain more than \(80\%\)
low-divergence tokens.  Clipping and missing-answer-marker rates are small.
Most importantly, we separate standalone teacher capability from teacher
behavior on student prefixes.  Qwen3-32B solves 1,033 of 4,997 eligible prompts
in all four independent attempts.  Repeating the eight-seed diagnostic on this
selected teacher-\(4/4\) subset raises student sequence accuracy to
\(86.91\%\), yet agreement-on-failure remains \(14.76\%\) of response-token
mass.  A teacher that reliably solves a prompt from scratch can still assign
locally compatible likelihoods along a student trajectory that fails.

Finally, we compare pure OPD, correct-only OPD, and a same-prompt contrastive
variant as controlled probes.  They use the available signals in three simple
ways: retain failed trajectories, mask them, or contrast them at the sequence
level.  None consistently reduces agreement-on-failure; their validation
ordering is checkpoint-sensitive, and transfer beyond the near-domain
distribution is mixed.  We relate this result to an indistinguishability
limitation: a tokenwise rule based only on local teacher divergence \(d_t\) and
the trajectory outcome \(z(y)\) treats tokens with the same observed pair
\((d_t,z(y))\) identically, even when they play different roles in the
reasoning process.  \Cref{sec:training} discusses the positional information
needed to break this ambiguity.  We therefore present the work as a diagnostic
rather than a new training method.

Our contributions are:
\begin{enumerate}[leftmargin=*]
  \item We formulate outcome-confounded local supervision and introduce a
  four-regime diagnostic that resolves pointwise OPD signals by the outcome of
  the trajectory on which they were observed.
  \item We provide eight-seed measurements on two student--teacher pairs, plus
  threshold, sequence, format, length, and teacher-capability controls showing
  that agreement-on-failure is a substantial and persistent regime in our
  setting.
  \item We use three training probes---imitate, filter, and coarse
    contrast---to test simple fixes.  Together with an indistinguishability
    argument, the results identify three sources of additional localization
    information beyond local divergence and a trajectory-level outcome scalar.
\end{enumerate}

%% file: sections/02_background.tex
\section{Setting: Dense Supervision on Student Prefixes}
\label{sec:background}

\paragraph{On-policy distillation.}
Let \(\pi_\theta\) denote the student and \(\pi_T\) the teacher.  Given a prompt
\(x\), the student samples a response \(y=(y_1,\ldots,y_L)\).  OPD then evaluates
the teacher on each student-visited prefix \(s_t=(x,y_{<t})\), producing
\(\pi_T(\cdot\mid s_t)\) without replacing the student's trajectory by a teacher
solution.  The implementation used here records the sampled-token log-ratio
\[
  r_t = \log \pi_\theta(y_t\mid s_t)-\log \pi_T(y_t\mid s_t),
\]
whose expectation under \(y_t\sim\pi_\theta(\cdot\mid s_t)\) is the local
reverse KL.  This setup reduces the exposure mismatch of sequence distillation
on fixed teacher traces \citep{kim2016sequence,gu2023minillm}, but it does not
ensure that a student-visited prefix is recoverable or on a correct reasoning
path.

\paragraph{Reasoning outcomes.}
For mathematical reasoning, a rule-based verifier assigns a trajectory label
\(z(y)\in\{0,1\}\) from the final answer.  The dense OPD signal and this outcome
operate at different granularities.  The teacher can agree with a locally
plausible continuation after the student has already entered a failing path; it
can also disagree with a token on a correct alternative derivation.  Our goal is
to measure this local/global mismatch without treating either signal as a
direct token-level correctness label.

\paragraph{Experimental setting.}
The primary study uses Qwen3-8B as the student and Qwen3-32B as the teacher on a
5,000-prompt \texttt{math\_dapo} discovery split.  For each of eight independent
data seeds, we sample 32 prompts and four student responses per prompt, yielding
128 rollouts per seed and 1,024 in total.  Sampling uses temperature \(1.0\),
top-\(p=1\), repetition penalty \(1.05\), no thinking mode, and a maximum of
4,096 response tokens.  Teacher likelihoods are computed on the sampled student
tokens.  The one-step jobs reuse the training infrastructure, but all discovery
statistics are measured on the pre-update rollout batch.  We report pooled
token counts for the primary estimate and per-seed variation as a descriptive
stability check.  Additional protocol and implementation details appear in
\cref{app:details}.

%% file: sections/03_diagnostic.tex
\section{Outcome-Resolved Local Supervision}
\label{sec:diagnostic}

\IfFileExists{figures/fig1_diagnostic_lens.pdf}{%
\begin{figure*}[t]
  \centering
  \includegraphics[width=\textwidth]{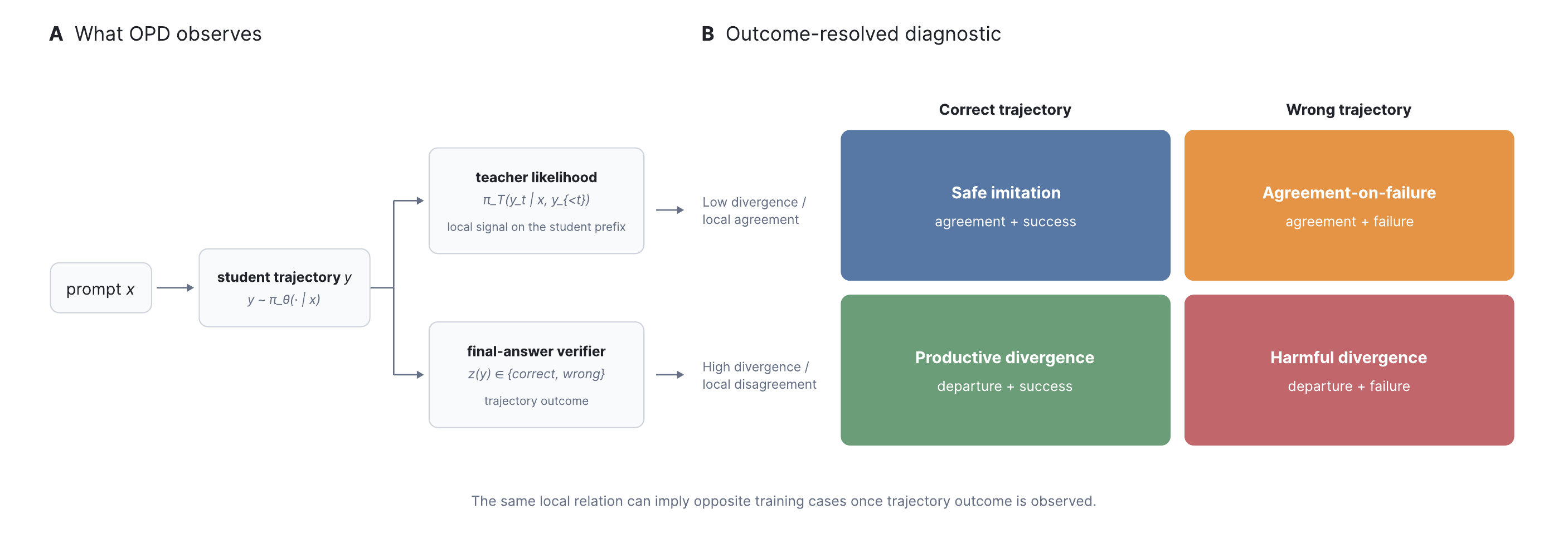}
  \caption{\textbf{Outcome-resolved view of local OPD supervision.}
  \textbf{A:} The teacher scores tokens at student-visited prefixes, while a
  final-answer verifier labels the same trajectory.
  \textbf{B:} Teacher--student divergence determines the rows and verifier
  outcome determines the columns, yielding four distinct regimes.}
  \label{fig:diagnostic_lens}
\end{figure*}
}{}

\paragraph{A pointwise divergence score.}
Individual samples of \(r_t\) can be negative, so we use the non-negative
\(k_3\) estimator
\[
  d_t = \exp(-r_t)-1+r_t
\]
for diagnostic binning.  Under student sampling, \(d_t\) and \(r_t\) have the
same reverse-KL expectation because
\(\mathbb{E}_{\pi_\theta}[\exp(-r_t)]=1\), while \(d_t\ge 0\) pointwise.  A
token is high-divergence when \(d_t>\tau\) and low-divergence otherwise; the
default is \(\tau=0.15\).  This is a pointwise diagnostic on the sampled token,
used only for binning; it does not identify the causal error in the trajectory.

\paragraph{Four regimes.}
Broadcasting the final outcome \(z(y)\) to the valid response tokens gives
\[
\begin{array}{c|cc}
 & z(y)=1 & z(y)=0\\ \hline
d_t\le\tau & \text{safe imitation} & \text{agreement-on-failure}\\
d_t>\tau & \text{productive divergence} & \text{harmful divergence}.
\end{array}
\]
The labels are descriptive, not normative token annotations.  Productive
divergence means only that a high-divergence token occurred on a successful
trajectory; it need not be the cause of success.  Likewise,
agreement-on-failure identifies low local divergence on a failed trajectory
without claiming that every such token is individually erroneous.

\paragraph{Aggregation.}
Our primary statistic is response-token mass because the OPD objective applies a
dense update over response tokens.  For example,
\[
  M_{\mathrm{AoF}} =
  \frac{\sum_{y,t}\mathbf{1}[z(y)=0]\mathbf{1}[d_t\le\tau]}
       {\sum_{y,t}\mathbf{1}[t\text{ is a valid response token}]}.
\]
We pool exact token numerators and denominators across seeds and separately
report seed variation and sequence-level checks.  This distinction is important:
wrong responses are longer on average, so token mass and sequence counts answer
related but non-identical questions.

%% file: sections/04_discovery.tex
\section{Discovery: Agreement Can Accompany Failure}
\label{sec:discovery}

\newcommand{\STwoSafe}{26.71}
\newcommand{\STwoProd}{1.19}
\newcommand{\STwoAoF}{67.68}
\newcommand{\STwoHarm}{4.43}
\newcommand{\STwoAcc}{32.13}
\newcommand{\STwoAoFmean}{67.81}
\newcommand{\STwoAoFstd}{3.38}

\begin{figure*}[t]
  \centering
  \vspace{-0.7em}
  \IfFileExists{figures/fig2_main_discovery.pdf}{%
    \includegraphics[width=\textwidth]{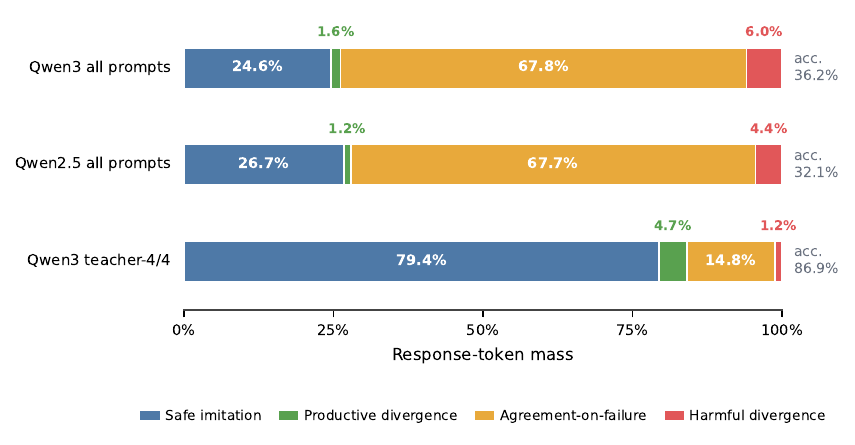}%
  }{%
    \figuremock{1.65in}{Figure 2}{fig2\_main\_discovery.pdf}%
  }
  \vspace{-0.6em}
  \caption{\textbf{Outcome-resolved response-token mass at
  \(\tau=0.15\).}  Each horizontal bar partitions pooled token mass into the
  four regimes.  Agreement-on-failure dominates the full eight-seed discovery
  distribution (top).  The second row repeats the same diagnostic on a
  Qwen2.5-7B/32B pair.  On the selected Qwen3 teacher-\(4/4\) subset
  (bottom), student accuracy is much higher, yet all four regimes remain.
  Cross-row shifts are descriptive because the teacher-\(4/4\) subset is
  easier and selected by teacher performance.}
  \label{fig:main}
\end{figure*}

\paragraph{Agreement-on-failure dominates token mass.}
The top row of \cref{fig:main} reports the eight-seed aggregate.  Safe imitation accounts
for \(24.58\%\) of valid response tokens, productive divergence for
\(1.61\%\), agreement-on-failure for \(67.84\%\), and harmful divergence for
\(5.96\%\).  The largest cell therefore contains locally low-divergence tokens
from responses that end incorrectly, not tokens marked by conspicuous
student--teacher disagreement.  The pattern is stable across seeds: the
unweighted seed mean for agreement-on-failure is \(67.90\%\), with a standard
deviation of \(4.83\) percentage points.

\paragraph{Disagreement is outcome-confounded as well.}
High-divergence tokens occupy \(7.58\%\) of pooled token mass, but they split
between successful and failed trajectories.  Harmful divergence is larger
than productive divergence (\(5.96\%\) versus \(1.61\%\)), so uniformly
rewarding divergence would mix successful and failed cases.  Productive
divergence is nevertheless nonzero in every seed, so treating every local
departure as an error is also unsupported.  At the token level,
\(P(\text{correct}\mid\text{high divergence})=21.28\%\), compared with
\(P(\text{correct}\mid\text{low divergence})=26.60\%\).  In this dataset,
pointwise divergence alone is therefore a weak classifier of trajectory
success.

\paragraph{External validity: a second student--teacher pair.}
To test whether agreement-on-failure is specific to the Qwen3-8B/Qwen3-32B
pair, we repeat the eight-seed discovery diagnostic without modification on a
second pair---a Qwen2.5-7B-Instruct student and a Qwen2.5-32B-Instruct
teacher---reusing the same prompts, protocol, verifier, and threshold
\(\tau=0.15\); only the models change.  The qualitative pattern persists.  Safe
imitation accounts for \STwoSafe\% of valid response tokens, productive
divergence for \STwoProd\%, agreement-on-failure for \STwoAoF\%, and harmful
divergence for \STwoHarm\%, at a student sequence accuracy of \STwoAcc\%.
Agreement-on-failure again occupies the largest cell, with an unweighted seed
mean of \STwoAoFmean\% (standard deviation \STwoAoFstd{} points).  The regime is
therefore not specific to the original model pair.

\paragraph{Scope of the claim.}
The four cells partition dense OPD signals; they do not localize the causal
reasoning error.  The empirical claim is that the same local relation occurs
under opposite trajectory outcomes at nontrivial scale.  Consequently, a rule
that consumes local agreement or disagreement without outcome context merges
qualitatively different training cases.

%% file: sections/05_robustness.tex
\section{Stress Tests and Teacher-Capability Control}
\label{sec:robustness}

\begin{figure*}[t]
  \centering
  \IfFileExists{figures/fig3_threshold_robustness.pdf}{%
    \includegraphics[width=\textwidth]{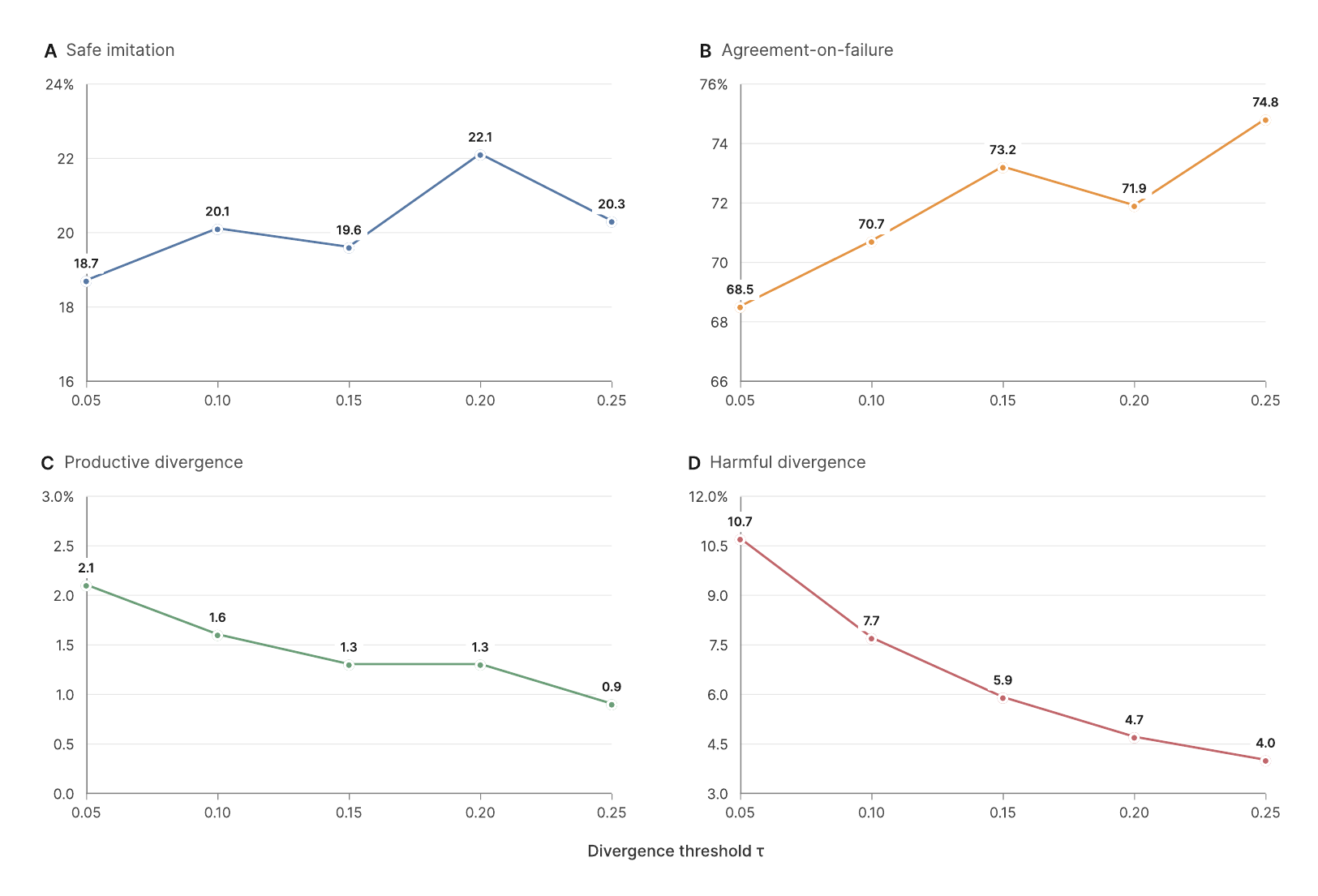}%
  }{%
    \figuremock{2.20in}{Figure 3}{fig3\_threshold\_robustness.pdf}%
  }
  \caption{\textbf{Threshold robustness in matched three-seed reruns.}
  Varying \(\tau\) from \(0.05\) to \(0.25\) changes the absolute partition as
  expected, but agreement-on-failure remains large and both high-divergence
  regimes remain nonzero.  Points pool exact token counts over three seeds at
  each threshold.}
  \label{fig:threshold_robustness}
\end{figure*}

\paragraph{Threshold sensitivity.}
\Cref{fig:threshold_robustness} sweeps
\(\tau\in\{0.05,0.10,0.15,0.20,0.25\}\).  Agreement-on-failure remains between
\(68.54\%\) and \(74.82\%\) of response-token mass.  Productive divergence
remains nonzero (\(0.90\%\)--\(2.09\%\)), as does harmful divergence
(\(4.02\%\)--\(10.70\%\)).  These reruns use three matched data seeds per
threshold; their purpose is qualitative sensitivity analysis rather than a
second estimate of the eight-seed default partition.

\begin{figure*}[t]
  \centering
  \IfFileExists{figures/fig4_sequence_sanity.pdf}{%
    \includegraphics[width=\textwidth]{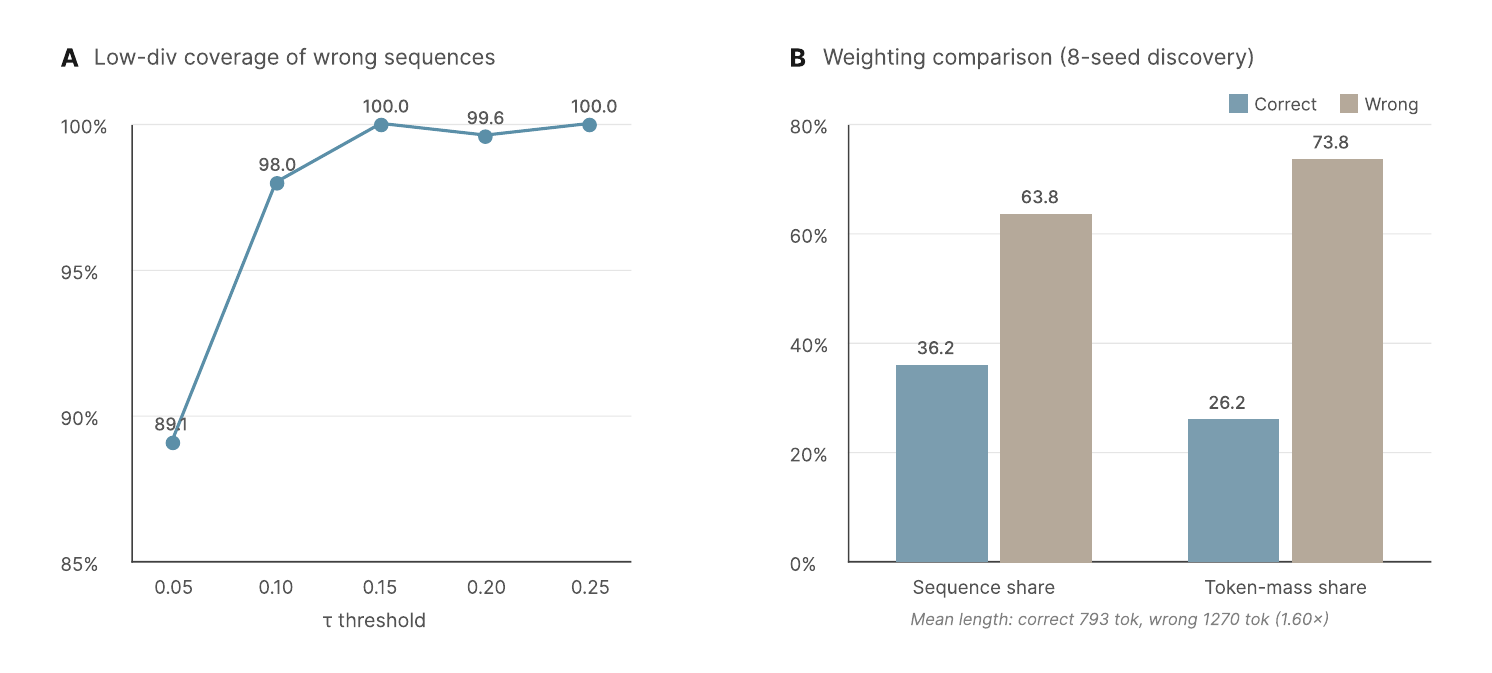}%
  }{%
    \figuremock{2.05in}{Figure 4}{fig4\_sequence\_sanity.pdf}%
  }
  \caption{\textbf{Sequence-level and token-level accounting.}
  \textbf{A:} In the three-seed threshold reruns, most wrong trajectories are
  themselves dominated by low-divergence tokens.
  \textbf{B:} In the default eight-seed discovery run, longer wrong responses
  receive more weight under token-mass accounting than under sequence
  accounting.  The two panels address complementary, not identical, audits.}
  \label{fig:sequence_sanity}
\end{figure*}

\paragraph{Token versus sequence accounting.}
In the default eight-seed discovery run, sequence accuracy is \(36.23\%\), but
correct trajectories contribute only \(26.20\%\) of token mass.  Correct
responses contain \(793.2\) valid tokens on average, compared with \(1{,}269.7\)
for wrong responses.  Token mass is the quantity directly exposed to a dense
OPD objective, but this \(1.60\times\) length ratio motivates an explicit
sequence-level check.  Across the threshold reruns,
\(89.15\%\)--\(100\%\) of wrong trajectories contain more than \(80\%\)
low-divergence tokens (\cref{fig:sequence_sanity}A).  Thus the effect is visible
within individual wrong trajectories and is not solely an aggregate
length-weighting artifact.

\paragraph{Format and truncation.}
Only \(1.76\%\) of default discovery responses reach the response limit, and
\(0.68\%\) of dumped samples lack a final-answer marker.  No closing-think
markers are missing under the no-thinking protocol.  These rates are too small
to explain the reported quadrant masses, although the rule-based outcome
scorer remains an explicit limitation.

\FloatBarrier

\paragraph{Does the teacher simply not know the answer?}
We independently sample Qwen3-32B four times on 4,997 eligible discovery
prompts with the same decoding protocol and outcome scorer.  Per-generation
accuracy is \(38.15\%\), pass@4 is \(56.15\%\), and 1,033 prompts are solved in
all four attempts.  We then repeat the eight-seed diagnostic using only this
teacher-\(4/4\) subset.  As \cref{tab:teacher_control} and the lower row of
\cref{fig:main} show, the selected subset is substantially easier: student
sequence accuracy rises from \(36.23\%\) to \(86.91\%\).  Nevertheless,
agreement-on-failure remains \(14.76\%\) of response-token mass
(\(14.47\%\pm4.25\%\) across seeds), and every logged wrong sequence in the
pooled subset is dominated by low-divergence tokens.

\begin{table}[!ht]
\centering
\small
\caption{Teacher-capability control at \(\tau=0.15\).  Values are percentages.
The teacher-\(4/4\) subset is selected and easier, so cross-column differences
are descriptive rather than causal.}
\label{tab:teacher_control}
\begin{tabular}{@{}l
d{2.2}
d{2.2}@{}}
\toprule
Statistic
& \multicolumn{1}{c}{All prompts}
& \multicolumn{1}{c}{Teacher \(4/4\)} \\
\midrule
Safe imitation          & 24.58 & 79.43 \\
Productive divergence   &  1.61 &  4.66 \\
Agreement-on-failure    & 67.84 & 14.76 \\
Harmful divergence      &  5.96 &  1.16 \\
Sequence accuracy       & 36.23 & 86.91 \\
\bottomrule
\end{tabular}
\end{table}

This control separates prompt-level teacher competence from the usefulness of
teacher likelihoods on arbitrary student prefixes.  Reliable standalone
solvability therefore does not guarantee a corrective local signal after the
student enters a failing trajectory.

%% file: sections/06_training_dynamics.tex
\section{Controlled Training Probes}
\label{sec:training}

The diagnostic suggests that neither unconditional imitation nor simple local
filtering should be expected to repair wrong trajectories.  We test this
implication with three matched 150-step probes on a teacher-filtered training
set: pure OPD; correct-only OPD, which masks the distillation loss on wrong
responses; and correct-only OPD with a same-prompt contrastive regularizer
(SPC, \(\beta=0.01\)).  SPC raises the mean student log-likelihood of correct
rollouts relative to wrong rollouts for prompts that produce both outcomes.
The three probes cover three simple uses of the available signals: retaining
failed trajectories, masking them, or contrasting them at the sequence level.
They are diagnostic interventions rather than proposed final methods.

\begin{figure*}[!ht]
  \centering
  \IfFileExists{figures/fig5_training_probes.pdf}{%
    \includegraphics[width=\textwidth]{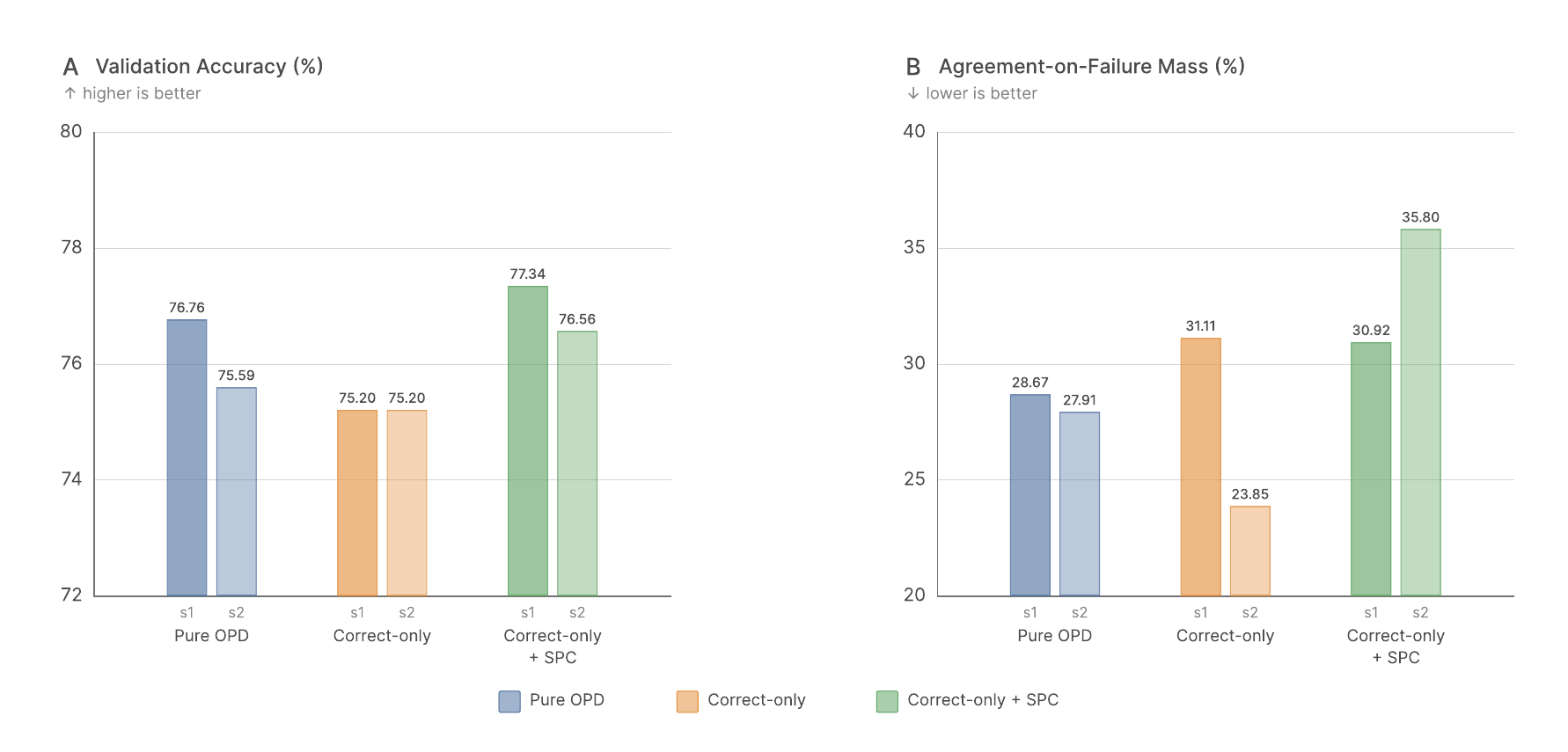}%
  }{%
    \figuremock{2.05in}{Figure 5}{fig5\_training\_probes.pdf}%
  }
  \caption{\textbf{Two-seed controlled probes at step 150.}  Within each
  method, the two labeled bars are the matched seeds.
  \textbf{A:} SPC has a small final-validation advantage.
  \textbf{B:} That advantage does not coincide with lower
  agreement-on-failure.  With only two seeds, the figure is diagnostic rather
  than a method-ranking result.}
  \label{fig:probes}
\end{figure*}

\paragraph{Small performance differences, persistent confounding.}
Across two seeds, final validation accuracy averages \(76.17\%\) for pure OPD,
\(75.20\%\) for correct-only, and \(76.95\%\) for SPC
(\cref{fig:probes}A).  Best-checkpoint means are \(78.91\%\), \(76.76\%\), and
\(78.22\%\), respectively, so the ordering changes under checkpoint selection.
More importantly, final agreement-on-failure averages \(28.29\%\) for pure
OPD, \(27.48\%\) for correct-only, and \(33.36\%\) for SPC
(\cref{fig:probes}B).  SPC improves final validation by \(0.59\) and \(0.98\)
percentage points relative to pure OPD on the two seeds, while increasing
agreement-on-failure on both.  A separate signed wrong-trajectory weighting
pilot also degrades by its final checkpoint.  These observations do not support
a claim that any probe solves the diagnosed failure mode.

\paragraph{Why the available signals cannot localize.}
The three probes share a localization limitation.  Consider a tokenwise
intervention of the form
\[
  w_t = f(d_t,z(y)),
\]
where \(d_t\) is the local teacher--student divergence and \(z(y)\) is the
trajectory outcome.  If two tokens share the same pair
\((d_t,z(y))\), then every such intervention must assign them the same
treatment, even if one token precedes the first reasoning error and the other
lies after the trajectory has become unrecoverable.  Thus \(f\) cannot
localize a corrective target inside a failed trajectory unless some additional
signal breaks this equivalence class.

The measurements show that this ambiguity occurs frequently.  First, \(d_t\)
is \emph{outcome-blind}:
\(P(\text{correct}\mid\text{low divergence})=26.60\%\) is close to
\(P(\text{correct}\mid\text{high divergence})=21.28\%\), and
\(89.15\%\)--\(100\%\) of wrong trajectories are themselves dominated by
low-divergence tokens, so \(d_t\) does not indicate which token turned the
trajectory wrong.  Second, \(z(y)\) is \emph{position-blind}: it is one scalar
broadcast identically to every response token.  Consequently, a locally valid
token before the first error and a token observed after the trajectory is
already lost can carry the identical signature
\((d_t\text{ low},\, z=0)\).  Imitation keeps both, filtering removes both
while supplying no corrective target, and coarse contrast repels both; the
persistent agreement-on-failure across the three probes is consistent with
this limitation.

\paragraph{Benchmark context.}
All six trained checkpoints improve the near-domain
\texttt{math\_dapo/test} set by \(3.64\)--\(5.28\) percentage points relative
to the base model.  MATH-500 changes range from \(+0.20\) to \(-1.00\) points,
and all MATH-L5 changes are non-positive.  AIME2025 contains only 30 prompts
and produces mixed greedy and pass@16 differences.  These results are too
limited to support broad-transfer or state-of-the-art claims.

\paragraph{Additional information required for localization.}
Repairing wrong trajectories requires positional information that \(d_t\) and
\(z(y)\) do not provide.  Three classes of additional signal could supply it:
a per-step correctness label (process supervision); a per-prefix recoverability
signal obtained by continuing the teacher from the student prefix \(s_t\)
(counterfactual teacher rollouts); or an alignment between correct and wrong
rollouts of the same prompt that marks their divergence point (token-level
cross-rollout localization).  Our SPC probe is a sequence-level version of the
third route: it compares whole-trajectory log-likelihoods without identifying
where the trajectories diverge, consistent with its failure to reduce
agreement-on-failure.  \Cref{sec:related} relates existing remedies, including
early rollout termination and process rewards, to these three sources of
localization information.

%% file: sections/07_exploratory.tex
\section{Discussion: What the Diagnostic Changes}
\label{sec:implications}

Agreement and disagreement alone do not determine credit assignment.
Outcome-resolved measurement changes how several common OPD operations should
be interpreted.

\paragraph{Filtering.}
Removing all wrong trajectories prevents imitation of agreement-on-failure, but
also discards information about where the student fails and provides no
replacement target.  Our correct-only probe is therefore a useful baseline,
not evidence that outcome filtering is sufficient.

\paragraph{Divergence weighting.}
Uniformly emphasizing high-divergence tokens mixes productive and harmful
divergence.  Uniformly suppressing them can penalize correct alternative
derivations.  A divergence threshold is therefore useful for measurement but is
not, by itself, a target policy.

\paragraph{Teacher strength.}
Standalone teacher accuracy and local teacher usefulness are different
properties.  Teacher filtering improves the prompt distribution, yet cannot
guarantee a corrective likelihood signal at every student-visited prefix.

\paragraph{Evaluation.}
Reporting only downstream accuracy hides whether an OPD objective improves by
expanding safe imitation, preserving productive divergence, reducing harmful
divergence, or merely changing response length.  The four-regime decomposition
provides a compact internal measurement layer to accompany performance curves.
It should not replace downstream evaluation, but it can reveal when similar
accuracy is reached through different supervision regimes.

\paragraph{Sources of localization information.}
The diagnostic identifies missing information rather than a single remedy.
Each route provides localization but requires additional supervision or
computation.  Process supervision provides per-step labels but presumes a step
verifier that the final-answer scorer does not; counterfactual teacher rollouts
from \(s_t\) give a per-prefix recoverability signal at the cost of additional
teacher generations; and token-level cross-rollout localization requires both
a correct and a wrong rollout for the same prompt plus an alignment between
them.  All three introduce information absent from purely local OPD.  Dense
teacher likelihoods alone do not provide positional, outcome-aware
supervision.  This also limits the scope of our negative training result: we do
not claim that agreement-on-failure is unfixable, only that rules using local
divergence and a trajectory outcome cannot localize the error without an
additional signal that breaks this ambiguity.

%% file: sections/08_related_work.tex
\section{Related Work}
\label{sec:related}

\begin{table}[t]
\centering
\scriptsize
\caption{\textbf{Positioning relative to OPD-adjacent work.}  The table marks
which ingredient is explicit in each line of work.  ``Outcome-resolved''
means that local teacher--student relations are split by the final outcome of
the same student trajectory; ``native-budget boundary'' means analyzing what
can and cannot be localized from \(d_t\) and \(z(y)\) alone.}
\label{tab:related_positioning}
\setlength{\tabcolsep}{3.0pt}
\begin{tabular}{@{}lcccc@{}}
\toprule
Line of work & Student prefixes & Dense teacher signal
& Outcome-resolved & Native-budget boundary \\
\midrule
GKD / OPD & \checkmark & \checkmark & -- & -- \\
KAT / prefix termination & \checkmark & \checkmark & partial & -- \\
Multi-rollout OPD & \checkmark & mixed & \checkmark & -- \\
Process / verifier signals & mixed & -- & \checkmark & -- \\
Ours & \checkmark & \checkmark & \checkmark & \checkmark \\
\bottomrule
\end{tabular}
\end{table}

\paragraph{Distillation on generated sequences.}
Classical knowledge distillation matches a student to a teacher distribution
\citep{hinton2015distilling}; sequence-level distillation extends the idea to
autoregressive outputs \citep{kim2016sequence}.  MiniLLM studies reverse-KL
distillation for generative language models \citep{gu2023minillm}.  GKD/OPD
instead samples from the student and evaluates the teacher on student-generated
sequences, reducing train--inference prefix mismatch
\citep{agarwal2024opd}.  We analyze a different mismatch that remains after
this move: a local teacher likelihood on a student prefix need not identify the
outcome of the complete student trajectory.

\paragraph{OPD analyses and structured supervision.}
Recent surveys and empirical analyses organize OPD mechanisms, pathologies, and
practical design choices
\citep{song2026opdsurvey,li2026rethinkingopd,wang2026demystifyingopd}.
Closest to us, KAT identifies a low-KL agreement failure after the student
enters a degraded prefix and remedies it with early rollout termination
\citep{xin2026kat}.  We differ in claim and in scope.  First, KAT's termination
is one instance of the counterfactual-recoverability route in our taxonomy: it
imports information about whether a prefix is still on a good path.  We
characterize the broader sources of localization information and explain why
local reweighting alone cannot substitute for them.  Second, our
teacher-\(4/4\) control shows that the regime persists (\(14.76\%\) of token
mass) after filtering by prompt-level teacher competence.  This control does
not establish the recoverability of individual student prefixes, which KAT
targets.  Third, we supply outcome-resolved accounting that quantifies both
agreement and disagreement split by outcome across seeds.  Multi-teacher OPD allocates
student rollouts across specialized teachers \citep{ma2026mopd}, while
multi-rollout OPD constructs supervision from peer successes and failures
sampled for the same prompt \citep{yu2026multirolloutopd}---an instance of the
cross-rollout localization route, of which our SPC probe is the non-localizing,
sequence-level degenerate case.

\paragraph{Reasoning with verifiable outcomes.}
RL with verifiable rewards optimizes trajectory-level outcomes for
mathematical reasoning, including GRPO-based systems such as DeepSeekMath and
large-scale open implementations such as DAPO
\citep{shao2024deepseekmath,yu2025dapo}.  DeepSeek-R1 further demonstrates that
rule-based outcomes can drive reasoning behavior at scale
\citep{deepseekai2025deepseekr1}.  Process-oriented approaches such as PRIME
seek denser credit assignment from outcome labels \citep{cui2025prime}, an
instance of the process-supervision route in our taxonomy.  OPD
already supplies dense teacher information; our results show that density does
not by itself resolve whether a local update belongs to a successful or failed
trajectory, and that closing this gap requires the same positional,
outcome-aware information these methods add rather than a finer local weighting.

%% file: sections/09_limitations_conclusion.tex
\section{Limitations and Conclusion}
\label{sec:limitations}

\paragraph{Limitations.}
The primary evidence uses one mathematical-reasoning pipeline and a rule-based
final-answer scorer.  Although we include a second student--teacher pair as an
external-validity check, the reported masses should not be transferred
numerically to other model families, tasks, decoding regimes, or process
verifiers.  The primary Qwen3 discovery sample contains 1,024 rollouts from a
larger prompt pool, and seed variation is descriptive rather than a
population-level confidence interval.  Thresholding compresses a continuous
pointwise score, although the qualitative partition is stable over the tested
range.  More fundamentally, broadcasting a final outcome to every response
token does not identify the causal error or certify every token on a successful
trajectory.  The teacher-\(4/4\) control is selected and easier; it isolates
standalone prompt solvability but not teacher reliability on arbitrary corrupted
prefixes.  Finally, the training comparison uses two seeds and a small set of
hand-designed probes, so it should not be read as a mature method benchmark;
and our insufficiency argument is an indistinguishability boundary supported by
these probes and by the outcome-blindness of \(d_t\), not a formal impossibility
theorem.  The three probes cover broad uses of the available signals but not
all possible objectives or hyperparameters, and we have not empirically
evaluated the three sources of localization information we identify.

\paragraph{Conclusion.}
Token-level teacher likelihoods do not by themselves identify which parts of a
trajectory should be reinforced or corrected.  Local teacher--student
agreement occurs on both correct and failed trajectories, and local
disagreement can be productive or harmful.  In our setting,
agreement-on-failure dominates the original discovery distribution and remains
measurable even on prompts the teacher solves in all four standalone attempts.
Threshold, sequence, length, and format audits show that the pattern is not
explained by bookkeeping choices alone.  The three interventions tested here
---imitation, filtering, and coarse contrast---do not consistently reduce it.
Together, the results identify a localization limitation for rules based only
on local divergence and a trajectory outcome.  Overcoming this limitation
requires positional, outcome-aware information, such as process labels,
teacher continuations from student prefixes, or token-level alignment across
rollouts.  Outcome-resolved diagnostics can then test whether the added signal
reduces this ambiguity.

%% file: sections/a_appendix.tex
\section{Additional Experimental Details and Tables}

\subsection{Discovery protocol}
\label{app:details}
The primary discovery pool contains 5,000 \texttt{math\_dapo} prompts.  Each
of eight jobs starts from the same Qwen3-8B checkpoint, uses an independent
data seed, samples 32 prompts, and generates four responses per prompt.
Sampling uses temperature \(1.0\), top-\(p=1\), top-\(k=-1\), repetition
penalty \(1.05\), no thinking mode, a maximum prompt length of 1,024 tokens,
and a maximum response length of 4,096 tokens.  Teacher likelihoods come from
Qwen3-32B evaluated on the student-generated prompt and response tokens.  The
outcome scorer is the rule-based \texttt{math\_dapo} final-answer scorer used
during training.  The jobs perform one update step for infrastructure reuse,
but all reported discovery statistics are computed on the rollout generated
before that update.

The default estimate pools exact token numerators and denominators over the
eight seeds.  Per-seed means and standard deviations are unweighted descriptive
summaries.  The threshold sweep uses three matched data seeds at each value of
\(\tau\), with exact per-sequence divergence fractions logged for the
sequence-level audit.

\subsection{Teacher-capability control}
The capability control independently generates four Qwen3-32B responses for
every eligible discovery prompt with the same decoding parameters.  Three
prompts exceed the shared prompt limit, leaving 4,997 completely scored prompt
groups.  The selected subset contains the 1,033 prompts for which all four
teacher responses are correct.  The matched diagnostic repeats the original
eight-seed protocol on this subset.  A prompt-concurrency cap is used only to
avoid inference-engine timeouts on long generations; it does not alter the
prompts, sampling parameters, or number of responses.

\subsection{Exact counts and seed-bootstrap intervals}
\label{app:ci_counts}
For the paper-facing ratios we also retain the exact logged numerators and
denominators.  \Cref{tab:appendix_ci_counts} reports a compact subset of these
counts together with descriptive seed-bootstrap intervals.  Each bootstrap
sample resamples the eight independent discovery jobs with replacement and
recomputes the pooled count ratio; it is therefore a seed-level uncertainty
summary rather than a token-i.i.d. confidence interval.

\begin{table}[ht]
\centering
\scriptsize
\caption{Exact logged counts and seed-bootstrap intervals for the core discovery statistics. Percentages are pooled count ratios.  Intervals resample independent seed jobs, not individual tokens.}
\label{tab:appendix_ci_counts}
\setlength{\tabcolsep}{3.2pt}
\begin{tabular}{@{}llrrrr@{}}
\toprule
Setting & Statistic & Numerator & Denominator & Pooled \% & 95\% CI \\
\midrule
Qwen3 all prompts & Safe imitation       & 34{,}520.4 & 140{,}422.9 & 24.58 & [21.46, 28.64] \\
Qwen3 all prompts & Productive divergence&  2{,}263.6 & 140{,}422.9 &  1.61 & [1.34, 1.89] \\
Qwen3 all prompts & Agreement-on-failure & 95{,}265.0 & 140{,}422.9 & 67.84 & [64.05, 70.91] \\
Qwen3 all prompts & Harmful divergence   &  8{,}373.9 & 140{,}422.9 &  5.96 & [5.35, 6.43] \\
Qwen3 all prompts & Sequence accuracy    &     46.4  &     128.0   & 36.23 & [33.11, 39.26] \\
Qwen3 all prompts & \(P(\mathrm{correct}\mid d_t>\tau)\) & 2{,}263.6 & 10{,}637.5 & 21.28 & [17.60, 25.51] \\
Qwen3 all prompts & \(P(\mathrm{correct}\mid d_t\le\tau)\)&34{,}520.4 &129{,}785.4 & 26.60 & [23.23, 30.90] \\
\midrule
Qwen3 teacher \(4/4\) & Agreement-on-failure & 12{,}600.5 & 85{,}360.0 & 14.76 & [11.80, 17.75] \\
Qwen3 teacher \(4/4\) & Sequence accuracy    &    111.3  &    128.0   & 86.91 & [84.77, 88.57] \\
\midrule
Qwen2.5 pair & Safe imitation       & 45{,}369.3 & 169{,}874.8 & 26.71 & [24.21, 29.11] \\
Qwen2.5 pair & Productive divergence&  2{,}025.0 & 169{,}874.8 &  1.19 & [1.08, 1.30] \\
Qwen2.5 pair & Agreement-on-failure &114{,}963.5 & 169{,}874.8 & 67.68 & [65.31, 70.06] \\
Qwen2.5 pair & Harmful divergence   &  7{,}517.0 & 169{,}874.8 &  4.43 & [4.11, 4.77] \\
Qwen2.5 pair & Sequence accuracy    &     82.3  &     256.0   & 32.13 & [29.59, 34.38] \\
\bottomrule
\end{tabular}
\end{table}

\subsection{Length, format, and sequence audits}
The default discovery audit covers 1,024 decoded responses.  Sequence accuracy
is \(36.23\%\), whereas correct responses account for \(26.20\%\) of token
mass.  Correct responses contain \(793.19\) valid tokens on average and wrong
responses contain \(1{,}269.70\), a \(1.601\times\) ratio.  Response clipping
is \(1.76\%\); \(0.68\%\) of dumped responses lack a final-answer marker, and
no closing-think markers are missing under the no-thinking protocol.  In the
threshold reruns, \(89.15\%\)--\(100\%\) of wrong trajectories contain more
than \(80\%\) low-divergence tokens.

\subsection{Illustrative agreement-on-failure trace}
\label{app:case_trace}
\begin{figure}[ht]
  \centering
  \includegraphics[width=0.98\linewidth]{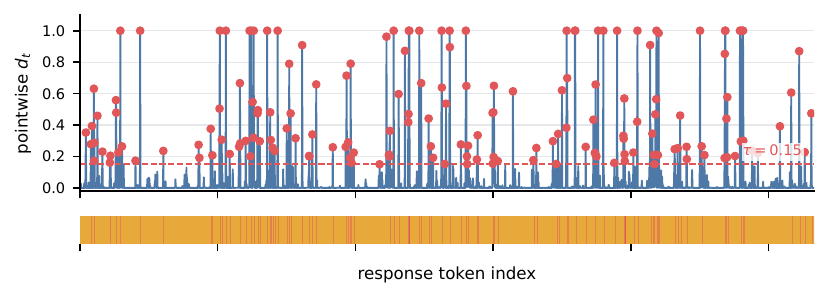}
  \caption{\textbf{Illustrative agreement-on-failure trace.}  A wrong response
  from the Qwen3 discovery run has final answer \(16\) while the gold answer is
  \(14\); \(93.7\%\) of its 2,666 response tokens are low-divergence at
  \(\tau=0.15\).  The top panel plots pointwise \(d_t\), clipped at 1.0 for
  readability; red markers indicate tokens above the threshold.  The bottom
  strip marks low-divergence tokens in orange and high-divergence tokens in red.}
  \label{fig:case_agreement_trap}
\end{figure}

\Cref{fig:case_agreement_trap} shows one traced wrong response from the primary
Qwen3 discovery run.  The student ultimately answers \(16\) while the gold
answer is \(14\).  We re-score the sampled response tokens with the Qwen3-8B
student and Qwen3-32B teacher and compute the same pointwise \(d_t\) diagnostic
used in the aggregate analysis.  The example is illustrative rather than a
causal token annotation: it shows that a failed trajectory can be locally
compatible with the teacher over most of its length.  In this response,
\(93.7\%\) of response tokens are low-divergence at \(\tau=0.15\).

\subsection{Controlled-probe protocol}
The two-seed comparison uses matched 150-step runs for pure OPD, correct-only
OPD, and correct-only OPD with SPC.  Each step samples eight responses for each
of 64 prompts at temperature \(1.2\), top-\(p=1\), and top-\(k=-1\), with no
thinking mode and a maximum response length of 8,192 tokens.  The actor
learning rate is \(5\times10^{-7}\); validation and checkpointing occur every
30 steps on a fixed 512-prompt development split.  Correct-only OPD sets the
distillation weight to zero for failed trajectories.

For every prompt that has both correct and wrong on-policy rollouts in the
current batch, SPC computes the mean length-normalized response log-probability
\(s^+\) over correct rollouts and \(s^-\) over wrong rollouts, then adds
\[
  \mathcal{L}_{\mathrm{SPC}}
  = -\log \sigma(s^+-s^-)
\]
with coefficient \(\beta=0.01\) to the correct-only OPD objective.  It uses no
teacher-repaired or off-policy response.  We report the final checkpoint for
the diagnostic comparison and separately retain the best development
checkpoint to expose checkpoint sensitivity.